\documentclass[12pt]{article}
\usepackage{lipsum} 				\usepackage[margin = 1in, left =1.5in,includefoot]{geometry}
\usepackage[hidelinks]{hyperref}		\usepackage{enumerate}

\usepackage{hyperref}
\hypersetup{
    colorlinks=true,
    linkcolor=blue,
    filecolor=magenta,      
    urlcolor=cyan,
}
\usepackage{listings}
\usepackage{graphicx}		
\usepackage{float}			
\usepackage{mhchem}		
\usepackage{fancyhdr}
\usepackage{algorithmicx}
\usepackage{algpseudocode}
\algdef{SE}[DOWHILE]{Do}{doWhile}{\algorithmicdo}[1]{\algorithmicwhile\ #1}%

\usepackage[english]{babel}
\usepackage[utf8]{inputenc}
\usepackage{amsmath}
\usepackage{amsfonts}
\usepackage{graphicx}
\usepackage[colorinlistoftodos]{todonotes}
\usepackage{algorithm}
\usepackage{algpseudocode}

\usepackage{geometry}
\begin{document}

\begin{titlepage}

\begin{center}
\baselineskip=14pt

\vspace*{5.7cm}
{\large\bf Scraping and Preprocessing Commercial Auction Data for Fraud Classification}\\
\baselineskip=14pt
\vspace*{6pt}

{\large Ahmad Alzahrani and Samira Sadaoui}\\ 
\vspace*{6pt}
{\large Technical Report CS 2018-05}\\ 
DOI: 10.6084/m9.figshare.6272342
\vspace*{4.5cm}

Copyright \copyright{} 2018, A.\ Alzahrani, S.\ Sadaoui\\ 
Department of Computer Science\\
University of Regina\\
Regina, SK, CANADA\\
S4S 0A2\\
\end{center}
\end{titlepage}

\thispagestyle{empty}		
\cleardoublepage 			
\pagenumbering{arabic}		
\setcounter{page}{1}

\begin{abstract}
In the last three decades, we have seen a significant increase in trading goods and services through online auctions. However, this business created an attractive environment for malicious moneymakers who can commit different types of fraud activities, such as Shill Bidding (SB). The latter is predominant across many auctions but this type of fraud is difficult to detect due to its similarity to normal bidding behaviour. The unavailability of SB datasets makes the development of SB detection and classification models burdensome. Furthermore, to implement efficient SB detection models, we should produce SB data from actual auctions of commercial sites. In this study, we first scraped a large number of eBay auctions of a popular product.  After preprocessing the raw auction data, we build a high quality SB dataset based on the most reliable SB strategies. The aim of our research is to share the preprocessed auction dataset as well as the SB training (unlabelled) dataset, thereby researchers can apply various machine learning techniques by using authentic data of auctions and fraud.  
\end{abstract}

\section{Introduction}
\subsection{Auction Fraud}
Online auctions have become a very lucrative e-commerce application. As an example, eBay, which is the largest auction site, recorded a net revenue of 9.7 billion U.S dollars in 2017, and the number of active users reached 170 million worldwide \footnote{https://www.statista.com}. Despite their popularity, e-auctions are very attractive to malicious moneymakers because auctions are vulnerable to cyber-crimes. This vulnerability is due to several facts, such as very low fees of auction services, anonymity of users, flexibility of bidding, and restricted legal policies.  According to the Internet Crime Complain Centre (IC3), auction fraud represents one of the top cyber-crimes.  For instance, in 2016, the count of auction complaints in only three states, California, Florida and New York, reached 7,448, and the victims' financial loss increased to \$5,900,977 \cite{CI32016}.  Auction users can commit different kinds of fraud, which are classified w. r. t. the time periods in which fraudulent activities can happen: 
\begin{itemize}
\item Pre-auction fraud, such as misrepresentation of products, auctioning of black market merchandise and stolen products.
\item In-auction fraud, which happens during the bidding period, such as shill bidding, bid sniping and bid shielding.
\item Post-auction fraud, such as non-delivery of products,  product insurance and fees stacking.
\end{itemize}

Both pre- and post-auction fraud can be noticed by users as it relies on concrete evidence. Nevertheless, in-auction fraud do not leave any clear evidence, and worst of all, it is not noticed by honest bidders and victims i.e. auction winners \cite{Kobayashi2015}.  Indeed, it is challenging to detect fraud occurring during the bidding period, and in our study, we are concerned about this kin of fraud.

\subsection{Shill Bidding Fraud}
Shill Bidding (SB) is the most common auction fraud but the most difficult to detect due to its similarity to normal bidding behaviour \cite{Dong2009, Ford2012}.  A shill bidder is a malicious user (the fraudulent seller and/or his accomplices) who bids aggressively in order to drive up the price of the product only to benefit the owner of the auction.  SB may cause a massive money loss for genuine sellers and bidders in the context of high priced products and also products with unknown value in the market, such as antiques \cite{BRANDLY2016}. As mentioned in \cite{trevathan2018, Dong2009}, excessive SB could lead to a market failure.  Online auctions may affect the users' confidence, which may negatively impacts the auctioning business.  In fact, several sellers and their accomplices have been prosecuted due to SB activities, including: 

\begin{itemize}
\item  In 2001, three sellers were charged of SB fraud worth a pay-off of \$300,000 through 1100 auctions of art paintings.  The fraud was conducted on eBay with more than 40 fake accounts \cite{trevathan2018}. 

\item In 2007, a jewellery seller was accused of conducting SB fraud on eBay, and had to pay \$400,000 for a settlement. Also, he and his employees were prevented from engaging in any online auctioning activities  for four years \footnote{https://www.nytimes.com/2007/06/09/business/09auction.html}.

\item In 2010, a seller faced a \pounds 50,000 fine after being found outbidding himself on eBay. He claimed that: ``{\it eBay let me open up the second account and I gave all my personal details and home address to do so.}"  \footnote{http://www.dailymail.co.uk/news/article-1267410/Ebay-seller-faces-fine-bidding-items-raise-prices.html}. 

\item In 2012, the online auction Trade Me had to pay \$70,000 for each victim after the investigation discovered SB fraud conducted by a motor vehicle trader in Auckland. The fraud was carried out for one year, and caused a significant loss for the victims. Trade Me blocked this trader from using their site, and referred the case to the Commerce Commission for a further investigation \footnote{https://www.trademe.co.nz/trust-safety/2012/9/29/shill-bidding}. 

\item In 2014, a lawsuit was filled against Auction.com by VRG in California claiming that the website allowed SB. The bid of \$5.4 million should have secured the property as the plaintiff declared, and yet the winning price was 2 million more. Auction.com was accused of helping the property’s loan holder, which is not fair for genuine bidders.  The California state passed a law on July 1, 2015, which requests the property auctioneers to reveal bids they submit on a seller’s behalf \footnote{https://nypost.com/2014/12/25/lawsuit-targets-googles-auction-com}. The spokeswoman for the California Association of Realtors said: ``{\it To the best of our knowledge, we’re the only state to pass this sort of legislation, even though we believe shill bidding to be prevalent all over the country.}" 
\end{itemize}

eBay policy states that ``{\it Shill bidding can happen regardless of whether the bidder knows the seller. However, when someone bidding on an item knows the seller, they might have information about the seller's item that other shoppers aren't aware of. This could create an unfair advantage, or cause another bidder to pay more than they should. We want to maintain a fair marketplace for all our users, and as such, shill bidding is prohibited on eBay}. \footnote{https://www.ebay.com/help/policies/selling-policies/selling-practices-policy/shill-bidding-policy?id=4353}". This statement clearly demonstrates that SB is troublesome and tough to be addressed.

\subsection{Generation of Shill Bidding Data}
Safeguard against SB fraud is lacking because of several issues, including:
\begin{itemize}
\item The  difficulty of extracting authentic data from auction websites.
\item The laboriousness of preprocessing the raw auction data.
\item The difficulty of identifying proper SB patterns.
\item The complexity of defining metrics for the SB patterns and measuring them from the original auction dataset.
\end{itemize}

To produce a reliable fraud training dataset,  SB patterns must be measured from original auction data. The latter represent the real behaviour of auction users, which is important to develop robust SB classification models and perform valid empirical assessment.  In commercial auctions, there is a tremendous amount of data that can be collected to provide useful information about the behaviour of users.  However, obtaining online auction data is a very tedious task as demonstrated in the related work section. In our study, we have employed a robust commercial Web scraper that is able to capture a large volume of data from completed auctions of a certain product.  Nevertheless, the original dataset contains irrelevant and redundant attributes, missing values and inappropriate value formatting. Therefore, preprocessing the auction dataset is a critical step before measuring the SB patterns and building robust fraud classifiers.  Still,  data preprocessing is a very time consuming phase as shown in this present report. The aim of this research is to share the preprocessed auction dataset as well as the SB training (unlabelled) dataset, thereby researchers can apply various machine learning techniques by using authentic auction and fraud data.

\section{Related Work}
Online auction sites, such as eBay, Taiwan, Yahoo!, TradeMe and uBid,  provide massive amounts of valuable data (made public) that can be utilized by researchers to understand the behaviour and trends of auction users as well as the popularity of products. Nevertheless, extracting data from websites and then structure them into a proper dataset are very challenging tasks.  To this end, researchers may employ commercial (general) Web scrapers, or implement their own auction scrapers.  But very few developers built their own auction scrapers to pursue their studies \cite{Yu2008,Balingit2009, Pandit2007}.  However, for some reasons that we did not come across, the software of the auction crawlers are not available for the public use. \\

Due to the complexity and cost of collecting auction data, developers introduced different techniques to improve the scraping efficiency.  For instance, \cite{Yu2008} implemented a parallel crawling tool based on multi-agent technology. Two types of crawling agents have been introduced: one to find users in the auctions and store them in a list of user IDs to avoid duplication; the other one to check the user ID list and crawl the unvisited users, and create the users' profile by capturing the auction contents. The experiment took place on eBay where only 7,682 pages were collected in an eight-hour time period, but without collecting the bidding list of each auction. If those lists were captured, the time may significantly increase. 

Another extraction tool is suggested in \cite{Balingit2009} where the completed auctions are extracted immediately while the ongoing auctions are first traced by the software, and once finished, data are collected. This tool is concurrently set up on several desktop computers to gather data using different search criteria of different auctions.   Even though it was running on multiple computers, the tool extracted only 1,300 auctions during a one-month period. On the other hand, \cite{Pandit2007} applied a queue technique to ensure that a user is not crawled more than once. Every seen user who has not been crawled yet is stored in the queue. For each loop, the first user is popped and all his feedback ratings are crawled. Then the popped user is marked as visited and stored in a different queue. To increase the crawling time efficiency, parallelism is engaged using a naive breadth-first search technique.  \\

Another option of collecting data is to utilize commercial Web crawlers, which provide portable software and services for users. The latter can purchase the license to download the scraper software, then extract the desired data. Also, users can have a contract with the web scraping provider to capture data for them and then deliver data in a certain format (CSV, Excel and API).  There are two main issues faced by any Web scraper, be it general or auction specific: 1) the developers must obtain permission to crawl data from the desired auction site; 2) after a certain period (usually 2 to 3 months), auction sites delete data because of the storage limitation  issue \cite{SadaouiSadaoui2017}. 

\section{Extraction of Raw In-Auction Data}
There are several professional web scrapers, such as Octoparse and Mozenda \cite{ahamad2017}, which extract data from any websites. We employed the fully automated scraping service provided by Octoparse \footnote{https://www.octoparse.com} that is able to effectively collect a large number of auctions of a certain product, including all the details about the auctions, bidders and bids. In the main page containing the auction listing, each product page is parsed and information are then extracted.  \\

We crawled from eBay the auctions of iPhone 7 for a period of three months, March to June 2017.  We selected this product for reasons that may increase the chance of SB activities: 
\begin{itemize}
\item It is in high demand since it attracted a high number of bidders. After tracing the eBay search results for all types of iPhones, we obtained a daily average of 3808 iPhones auctioned in June 2017 \footnote{http://www.ebay.com/sch/Cell-Phones-Accessories}. 

\item It is marked as a ``hot" product on eBay, which means it is among the most sold items in its category.  According to Terapeak  website \cite{terapeak2016},  93\% of iPhone 7 sales belong to the ``cellphones and accessories" category. 
 
\item It has a good price range with the average of \$610.17 (U.S currency). More the item price is high, more the possibility of fraud. Indeed, there is a direct relationship between SB activities and the auction price \cite{Sadaoui2017}.

\item The bidding duration varies between 1, 3, 5, 7 and 10 days.  In long auction duration, a shill bidder may easily mimic usual bidding behaviour \cite{Dong2009}. However, as claimed in \cite{chang2014}, fraudulent sellers receive positive rating in short bidding duration. Thus, we considered long and short bidding duration as shown in Table \ref{NO_Auctions_Duration}.
\end{itemize}

\begin{table}[H]
		\centering
			\caption{Number of Auctions for Each Duration}
			\begin{tabular}{|p{1.3in}|p{0.95in}|p{0.85in}|p{1in}|p{0.85in}|p{0.7in}|}
				\hline
				    {\bf  No. of Days} & 1 & 3 & 5 & 7 & 10 \\ \hline
                		{\bf No. of  Auctions} & 166 (20.57\%) & 187 (23.2\%) & 131 (16.23\%) & 309 (38.3\%) & 14 (1.7\%) \\							
               \hline							
			\end{tabular}
			\label{NO_Auctions_Duration}
		\end{table}

More precisely, we focused on the most popular auction protocol \cite{chen2003}: Forward (one seller and multiple buyers) and English (open ascending price bidding).  In addition, we utilized three filters to capture the iPhone 7 auctions: 1) from eBay.com of North America, 2) from ``cellphones and accessories" category, and 3) the wining price in each auction must be more than \$100.   In Table \ref{Statistics1}, we provide statistical information about the collected iPhone 7 auctions before the preprocessing task.  

\begin{table}[H]
		\centering
			\caption{Statistics of iPhone 7 Auctions}
			\begin{tabular}{|p{2.2in}|p{.9in}||p{1.6in}|}
				\hline
				    & {\bf Raw Data} & {\bf Preprocessed Data} \\ \hline
                	Number of Auctions & 2551 & 807 \\ \hline				
              		Number of Records & 399206 & 15145 \\ \hline					
                	Number of Bidder IDs & 1226 & 1054  \\ \hline
				 	Number of Seller IDs & 1727 & 647 \\ \hline
				 	Avg. Winning Price & \$ 610.17 & \$ 578.64  \\ \hline	
				 	Avg. Auction duration & 5 & 7  \\ \hline
					Number of attributes & 28 & 12  \\	
               		\hline							
			\end{tabular}
			\label{Statistics1}
		\end{table}
\section{ Preprocessing of In-Auction Data}
Original datasets often contain defective, missing and duplicated data that must be resolved to avoid misleading the learning process and returning an undesirable classification performance. Our preprocessing phase consists of transforming auction data into suitable data to be able to produce high quality SB data. More precisely, we conducted several preprocessing tasks, including data cleansing, re-formatting, aggregation and addition.

\subsection{Data Cleansing}
Firstly, we need to remove noisy data possessing the following characteristics: 
\begin{itemize}
\item Irrelevant and duplicated attributes: several attributes in the raw dataset are not needed to compute the SB metrics, such as the product location and ID,  feedback ratings of sellers and bidders, and bidders' account links. Also, some data are displayed twice on the main auction page and inside the auction link, such as the seller name, auction starting time, number of bids, and seller rating. There attributes are removed. 

\item Duplicated records: during the scrapping process, some data have been collected more than once, e.g., when a bidder participates more than once in an auction, the crawler collects his history each time. For example, let us suppose a bidder has 10 records in his bidding history,  and he participated two times in an auction, then the crawler will grab his history each time. As a result, we would receive 20 records of that bidder but 10 of them have been already captured.  

\item Records with missing values: there are several rows without the bidders' IDs; and we are not certain whether it was caused by eBay side or the scrapper itself. These IDs cannot be generated by using the inputting techniques. So, we need to delete them. 

\item Auctions with less than 5 bids: these auctions did not engage any SB fraud because the very few placed bids are genuine.  In some of these auctions, the items were sold using the ``Buy-It-Now" feature on eBay (bidders can pay off the price directly by clicking Buy-It-Now button). In some other auctions, sellers canceled their sales due to reasons like the items were not available anymore for sale, or error discovered in the listing \footnote{http://pages.ebay.co.uk/help/sell/questions/endlist-now.html}.

\item Auctions with inconsistent data:  several attributes contain incompatible values. For instance, the last submitted bid is greater than the winning price, or the starting price is greater than the winning price.  So, we decided to remove these auctions to not mislead the fraud classifiers. 
\end{itemize}

The statistics after cleaning noisy data are presented in Table 2.  As we can see, a good number of records have been deleted due to the issue of dupplicationg data.

\subsection{Attribute Aggregation}
We need to aggregate the date and time attributes into a single attribute for three features  as presented in Table \ref{Combine_date_time}. Also, we converted the month into a number. 

	\begin{table}[H]
		\centering
			\caption{Combining Date and Time into a Single Attribute}
			\begin{tabular}{|p{1.9in}|p{1.9in}||p{1.7in}|}
				\hline
                		{\bf Auction Starting Date} & {\bf Auction Starting Time} & {\bf Start Time} \\ \hline				
              		Jun-01-17 & 19:24:55 PDT & 06-01-17 19:24:55  \\ \hline					
                		{\bf Auction End Date} & {\bf Auction End Time} & {\bf End Time} \\ \hline				
              		Jun-03-17 & 19:24:55 PDT & 06-03-17 19:24:55  \\ \hline	
              		{\bf Bid Submission Date} & {\bf Bid Submission Time} & {\bf Bid Submit Time}\\ \hline				
              		Jun-02-17 & 08:20:50 PDT & 06-02-17 08:20:50  \\	
               		\hline							
			\end{tabular}
			\label{Combine_date_time}
		\end{table}

\subsection{Attribute Reformatting}
In Table \ref{Attributes_Reformatting}, four features, date, time, duration and price, must be reformatted into a quantitative value to be able to compute the SB metrics.  The date and time were both changed into seconds by calculating the seconds from the date we received the raw dataset (2017-07-07 00:00:00) to the date-time attributes in each auction.  Regarding the final price attribute, we deleted the currency and converted the value data-type from varchar to float. 

	\begin{table}[H]
		\centering
			\caption{Attribute Reformatting}
			\begin{tabular}{|p{1.9in}||p{1.9in}|}
				\hline	
                		{\bf Start Time} & {\bf Start Time Sec} \\ \hline				
              		2017-06-19 20:11:19 & 1741721  \\ \hline
              		{\bf End Time} & {\bf End Time Sec} \\ \hline				
              		2017-06-24 20:11:019 & 1309721  \\ \hline             		
              		{\bf Bid Submit Time} & {\bf Bid Submit Time Sec} \\ \hline				
              		2017-06-24 20:11:07 & 1309733  \\ \hline             		            		
             	{\bf Auction Duration} & {\bf Auction Duration Sec} \\ \hline				
              		5 & 432,000  \\ \hline             		
                		{\bf Winning Bid (varchar)} & {\bf Winning Bid (float)} \\ \hline				
              		650.50 \$ & 650.50  \\ 
               		\hline							
			\end{tabular}
			\label{Attributes_Reformatting}
		\end{table}
  
\subsection{Attribute Adjustement}
After a thourough examination of the whole auction dataset, we found out that several auctions contain attributes that are uncompatible with the existing list of bids. For example, the ``Number of Bids" attribute is different from the actual number of submitted bids in an auction. We have the same issue with the attribute "Number of bidders". We believe that these values have been corrupted during the crawling process. We fixed all the inconsistent values. 

\subsection{Attribute Addition}
In our original data, a specific URL represents an auction, which is not a proper format for measuring SB patterns.  To overcome this problem, a new attribute called AuctionID is given to provide a unique integer identifier for each auction.  However, the AuctionID is repetitive in the dataset w. r. t. number of bids in that auction. This will make the computation of the SB patterns time consuming.  So, the AuctionID cannot be the primary key for the records in the auction dataset. Thus, there is a need to represent several attributes: AuctionID, BidderID and Bid Submit Time Sec, with a single identifier for each record. Thus, a new attribute is added to the dataset called recordID to uniquely identify each individual record. The relevant set of auction attributes are presented in Table \ref{Auctions_Attributes}.

\begin{table}[H]
	\centering
		\caption{Auction Attributes for Shill Bidding}
			\begin{tabular}{|p{.15in}|p{2.25in}||p{3.4in}|}\hline
			\multicolumn{3}{|c|}{{\bf Auction Level}} \\
				\hline
					1 & Auction ID & Unique identifier of an auction \\ \hline
                	2 & Seller ID & Unique identifier of a seller   \\ \hline					
                	3 & Number of Bidders &  Number of bidders in an auction\\ \hline
                	4 & Starting Price  & Starting price set up by a seller  \\ \hline	
				 	5 & Auction Duration Sec &  How long an auction lasted  \\ \hline
				 	6 & Start Time Sec & Time on which an auction started\\ \hline
				 	7 & End Time Sec & Time on which  an auction ended\\ \hline
				 	\multicolumn{3}{|c|}{{\bf Bid Level}} \\\hline
				 	1 & Bidder ID &  Unique identifier of a bidder\\ \hline
				 	2 & Bid Amount & Bid price placed by a bidder   \\ \hline
				 	3 & Bid Submit Time Sec & Time where a bid was submitted by a bidder\\ \hline	
				 	4 & Number of Bids &  Number of submitted bids in an auction \\ \hline	
				 	5 & Winning Bid &  Final price of an auction  \\ \hline
					6 & Record ID & Unique identifier of a record in the dataset\\ 
				\hline							
		\end{tabular}
		\label{Auctions_Attributes}
\end{table}


\section{Characteristics and Metrics of Shil Bidding}
To increase the revenue of the seller, a shill bidder inflates the price of the product by  bidding aggressively. Once the price is high enough for the seller and the last submitted bid is of a honest bidder, the shill bidder stops competing, and thus making the normal bidder win the auction. Thus, the winning bidder pays much more than the real value of the product 

\subsection{Strategies of Shill Bidding}
By examining throughly the literature on the SB strategies \cite{Dong2010, Ford2012, Sadaoui2017}, we compiled below the most common and most relevant SB patterns. Moreover, the selected SB patterns are non redundant as they represent one unique aspect of the bidding behaviour. This is very important to build robust classifiers.

\begin{enumerate}
\item \textbf{Bidder Tendency:} a shill bidder participates exclusively in auctions of few sellers rather than a diversified lot.  This is a collusive act involving the fraudulent seller and an accomplice. The latter acts as a normal bidder to raise the price.

\item \textbf{Early Bidding:} a shill bidder tends to bid pretty early in the auction (less than 25\% of the auction duration) to get the attention of auction users.

\item \textbf{Bidding Ratio:} a shill bidder participates more frequently to raise the auction price and attract higher bids from legitimate participants.

\item \textbf{Last Bidding:} a shill bidder becomes inactive at the last stage of the auction (more than 90\% of the auction duration) to avoid winning the auction.

\item \textbf{Auction Starting Price:}  a shill bidder usually offers a small starting price to attract legitimate bidders into the auction.

\item \textbf{Successive Outbidding:} a shill bidder successively outbids himself even though he is the current winner to increase the price gradually with small consecutive increments. 

\item \textbf{Winning Ratio:}  a shill bidder competes in many auctions but hardly wins any auctions. 

\item \textbf{Auction Bids:} auctions with SB activities tend to have a much higher number of bids than the average of bids in concurrent auctions (i.e. selling the same product). Therefore, sellers of these auctions have a high probability of colluding with shill bidders to increase their profits. 
\end{enumerate}

We would like to mention that we did not use four other SB patterns proposed in the literature:
\begin{enumerate}
\item ``Nibble Bidding'' refers to a bidder who outbids others with a very small increment. We do not consider this pattern as a strong sign of fraud since normal bidders may do the same  \footnote{https://community.ebay.com/t5/Archive-Buyer-Central/newbie-buyer-Q-re-quot-nibbling-quot/td-p/2789578}. Additionally, this pattern is not present in the iphone7 auctions because the minimum bid increment has been fixed to \$5. 

\item ``Reserve Price Shilling" denotes a shill bidder who aggressively outbids himself as long as the current auction price is less than the reserved price.  We do not take into account this pattern for the simple reason that the reverse price is hidden by eBay. Nevertheless, Successive Outbidding and Bidding Ratio patterns are enough to cover this pattern.  

\item ``Buyer Rating" and ``Seller Rating" patterns represent the bidder's and seller's reputations in the auction house where buyers and seller can leave a feedback rating once the transaction has been completed.  We exculded these patterns due to the high potential of misusing this feature by fraudulent rings \cite{Ford2012,Dong2009}.  
\end{enumerate}

\subsection{Weights of Shill Bidding}
As presented in Table \ref{SB_Property_Weight}, we have categorized SB patterns in two dimensions: the first one illustrates the pattern category (bid-wise, bidder-wise or auction-wise), and the second one its relative weight.

\begin{enumerate}
\item Bidder property refers to the SB pattern that describes the user's participation in the whole auction dataset. These patterns are Bidder Tendency and Winning Ratio, which respectively examine how many times  a user attended to a specific seller, and his intention to whether win the auction or just inflate the price. 

\item Bid property illustrates  features related to the submitted bid, such as Early Bidding to check the time between the auction starting time and the first placed bid,  Bidding Ratio to compare the bidder's participation with other users, Successive Outbidding to check whether the bidder outbidded himself or not, and Last Bidding to verify whether a bidder became idle at the last stage of the auction. 

\item Auction property is inherent to the auction itself, such as Auction Starting Price, which is compared to the average of all the auctions' starting prices, and Auction Bids is compared to the average of the number of bids in all auctions. 
\end{enumerate}

As mentioned earlier, shill bidders may behave similarly to usual bidders \cite{SadaouiSadaoui2017, Ford2010}.  Each SB pattern reflects a specific aspect of the bidder's behaviour. Thus, it is necessary to study each pattern individually to provide a proper weight. But first let us consider the following two scenarios: (1) a bidder outbids others aggressively with the real intention of winning the auction, not to raise the item price. The problem with this scenario is that the bidding ratio is high;  (2) A seller gives low starting price just to attract more bidders in order to sell the product quickly. Therefore, to differentiate between suspicious and normal bidders, we assign different levels of weights: low, medium and high \cite{SadaouiSadaoui2017, Sadaoui2017}. 
\begin{enumerate}
\item \textbf{Low weight SB:}
Early Bidding and Auction Starting Price are given low weight because they occur very early in an auction.  The high number of auction bids might refer to the high quality of an item or the excellent reputation of a seller \cite{SadaouiSadaoui2017}. Thus, low weight is assigned to it as well. 
\item \textbf{Medium weight SB:}
In general, Last Bidding indicates the genuineness of the bidders to not win the auction \cite{Dong2010}. However, the cost of an item might get higher than what it is worth from the bidder's perspective, hence, he stops bidding. Therefore, medium weight is assigned to Last Bidding pattern. Medium weight is given to the Bidder Tendency pattern as well since on one side, some sellers make under table contract with bidders to increase their profits, but on the other side, the bidder is honest and desires to win the auction. Bidding Ratio is also considered of low weight since high bidding ratio might indicate the motive to win the auction.
\item \textbf{High weight SB:}
Since Winning Ratio pattern reflects the user behavior in the all auctions that he participated in, high weight is given to it. Also, the goal of shill bidders is not to win the auction but to increase the price of an item. A high weight is also assigned to Successive Outbidding pattern since normal users surely will not outbid themselves \cite{Sadaoui2017}. 
\end{enumerate}

As suggested in \cite{Sadaoui2017},  we manually assign the value of 0.3 to the low weight, 0.5 to the medium weight and 0.7 to the hight weight. 
\begin{table}[H]
		\centering
			\caption{Properties and Weights of SB Patterns}
			\begin{tabular}{|p{2in}|p{1in}|p{.6in}|p{.5in}|}
				\hline
				 \bfseries{ Fraud Pattern} & \bfseries{ Property}& \bfseries{Weight} & \bfseries{Value}\\ \hline
				 Bidder Tendency & Bidder & Medium & 0.5 \\ \hline
				Early Bidding & Bid & Low  & 0.3 \\ \hline
				Bidding Ratio & Bid & Medium  & 0.7 \\ \hline
				Last Bidding  & Bid & Medium & 0.5 \\ \hline
				Auction Starting Price& Auction & Low  & 0.3 \\ \hline 
				Successive Outbidding & Bid & High & 0.7 \\ \hline
				Winning Ratio & Bidder & High & 0.7 \\ \hline		
				Auction Bids & Auction & Low  & 0.3 \\ 				 			
				\hline							
			\end{tabular}
			\label{SB_Property_Weight}
		\end{table}

\subsection{SB Metrics}
The metrics are calculated from the auction dataset. Each metric is scaled to the range of [0, 1].  High values refer to a suspicious bidding behaviour.  Below $B$  denotes a bidder (Bidder ID) who competed in an auction $A$ (Auction ID) initiated by seller $S$ (Seller ID). The metrics  of the selected fraud patterns are provided below.

\begin{enumerate}

\item {\bf Bidder Tendency $(BT)$}

\begin{center} 

\begin{algorithmic}
 \State $BT(B,S) = 0$
  \If {$participateAllAuctions(B) > 1$}
    \State $BT(B,S) = \frac{participatewithSeller(B, S)}{participateAllAuctions(B)}$
  \EndIf
\end{algorithmic}

\end{center} 

where $participatewithSeller()$ denotes the total number of a bidder's participations for a specific seller, and $participateAllAuctions()$ the total number of auctions (from the iPhone 7 dataset) that a bidder participated in.
      

\item {\bf Early Bidding $(EB)$}

$$
   EB(B, A) = 1 - \frac{StartTimeSec_{A} - BidSubmitTimeSec_{B, A}}{AuctionDurationSec_{A}}
$$

\item {\bf Bidding Ratio $(BR)$}

$$
    BR(B, A) = \frac{totalBids(B, A)}{NumberofBids_{A}}
$$

where $totalBids()$ is the number of submitted bids by a bidder in an auction.
      	      	
\item {\bf Last Bidding $(LB)$}

$$
    	LB(B,A) = \frac{BidSubmitTimeSec_{B ,A} - EndTimeSec_{A}}{AuctionDurationSec_{A}}
$$

\item {\bf Auction Starting Price $(ASP)$}

\begin{center} 

\begin{algorithmic}
  \State $ASP(A) = 0$
  \If {$StartingPrice_{A} < avgAuctionsStartPrice(auctionDataset)$}
    \State $ASP(A) = 1 - \frac{StartingPrice_{A}}{avgAuctionsStartPrice(auctionDataset)}$
  \EndIf
\end{algorithmic}

\end{center} 
where $avgAuctionsStartPrice()$ is the average starting price in all the auctions from the dataset.


\item {\bf Successive Outbidding $(SOB)$}
\begin{center} 

    \begin{algorithmic}
    \State $SOB = 0$
     \Do             
     	\If {$(successiveBid(B, A) >= 4)$}
      	    \State $\{SOB(B, A) = 1;  quit \}$
        \ElsIf {$(successiveBid(B, A) >= 3)$}
      	     \State $\{SOB(B,A) = 0.5;  quit \}$
        \EndIf
     \doWhile {(notReached(lastBid(B, A))}
\end{algorithmic}

\end{center}

where $successiveBid()$ denotes the number of successive outbids submitted by a bidder $B$ in an auction $A$.


\item {\bf Winning Ratio $(WR)$}

$$
    	WR(B) =  
    1 - \frac{AuctionWon(B)}{auctionPartHigh(B)}
$$

$$
where ~~~~auctionPartHigh = count \{ A | BR (B,A) > 0.1\}
$$

where $AuctionWon()$ is the total number of auctions won by a bidder, and $auctionPartHigh()$ is the total number of auctions where the bidding ratio of the bidder is higher than 10\% of the total bids. This will eliminate the issue of non-active users. 


\item {\bf Auction Bids $(AB)$}

\begin{center} 

\begin{algorithmic}
    \State $AB(A) = 0$
  \If {$ NumberofBids_{A} > avgBidAllAuctions(auctionDataset)$}
    \State $AB(A)= 1 - \frac{avgBidAllAuctions(auctionDataset)}{NumberofBids_{A}}$
  \EndIf
\end{algorithmic}

\end{center}
where $avgBidAllAuctions()$ is the average number of bids in all the auctions in the dataset.

\end{enumerate}

\section{Generation of SB Data from In-Auction Data}
To produce SB data, the eight SB patterns are measured for all the bidders of the iPhone7 auctions. To accomplish this task, we use MSSQL to store the values of the 12 attributes for each auction, and then calculate each pattern against each bidder in each auction.   As a result, we generate a SB dataset with a tally of 6321 instances. An instance represents the conduct of bidders in a certain auction. It is a vector of 10 features: Auction ID, Bidder ID and the eight SB patterns.  Table \ref{statistics2} provides some statistical information about our SB dataset. We may mention that we examined carefully the whole SB dataset to check for outliers (values that are outside the range of [0, 1]), and we did not find any.  Outliers maybe present in a dataset due to corrupted data during the crawling process.
\begin{table}[H]
		\centering
			\caption{Statistics of SB Dataset}
			\begin{tabular}{|p{3.3in}|p{1in}|}
				\hline
                	Number of auctions & 807  \\ \hline				
              		Number of instances & 6321  \\ \hline					
                	Number of bidder IDs & 1054   \\ \hline
				 	Avg. bidders in low starting price auctions & 10   \\ \hline
					Avg. bidders in high starting price auctions & 6   \\ \hline	
					Winners and not aggressively participated & 536 (8.5\%)   \\ \hline	
					Not winners and aggressively participated & 1659 (26.25\%)  \\ \hline
					Not winners and not aggressively participated & 3817 (60.4\%)  \\ 
               		\hline							
			\end{tabular}
			\label{statistics2}
		\end{table}

From the SB dataset, we deduced the following facts as shown in Table \ref{statistics2}:

\begin{itemize}
\item 26.25\% of bidders aggressively outbidded themselves and others but did not win any auctions. The behavior of those bidders indicates that they committed SB.

\item 8.5\% of bidders did not highly participated in the auctions but won. Those bidders were fairly active at the last auction stage. All these refer to genuine behaviour.  

\item 60.4\% of bidders looked normal and did not win. Indeed, those bidders did not aggressively outbid others and did not submit successive bids.

\item 4.9\% of bidders extremely outbidded others and won the auctions. This indicates their desire to win the auction.

\item The average number of bidders in auctions with a  low starting price is 10, whereas the average is 6 in regular auctions with a regular starting price.

\item 11.5\% of bidders submitted bids at an early stage and aggressively outbidded others but did not win any auctions. This indicates their intention to increase the item price.
\end{itemize}

\begin{table}[H]
		\centering
			\caption{SB Patterns in the Auction Dataset}
			\begin{tabular}{|p{2in}|p{1.6in}|}
				\hline
					{\bf SB Pattern} & {\bf High Value ($>$ 0.7)}\\ \hline
                		Bidder Tendency & 209 (3.3\%) \\ \hline
                		Early Bidding & 2112 (33.4\%) \\ \hline					
                		Bidding Ratio & 43 (0.68\%) \\ \hline
                		Last Bidding & 1976 (31.26\%) \\ \hline
                		Auction Start Price & 2944 (46.6\%) \\ \hline
						Successive Outbidding & 1968 (31.13\%) \\ \hline
						Winning Ratio & 1659 (26.24\%) \\	\hline	
						Auctions Bids & 221 (3.5\%) \\ 
               	\hline							
			\end{tabular}
			\label{statistics3}
		\end{table}
		
As presented in Table \ref{statistics3}, Successive Outbidding pattern is relatively high since 31.13\%  of  samples/bidders  aggressively outbiddded others. Last Bidding pattern also shows that 31.26\% of bidders remained inactive at the last auction stage  whereas 33.4\% of bidders  participated at the early stage. A good number of bidders (46.6\%) tends to bid in auctions with a low starting price. 

\section{Future Work}
Here, we provide the future directions of our research work, which is illustrated in Figure 1:
\begin{itemize}
\item One of the most challenging tasks in machine learning is to label the training dataset. Subsequently, the next phase is to label the SB data into ``Normal" and ``Suspicious"  by using two main methods: unsupervised learning, such as clustering, or semi-supervised learning.  

\item Fraud data are highly imbalanced, and this class imbalance has been shown to reduce the performance of baseline classifiers as demonstrated in \cite{Sadaoui2018}. In this situation, the classifiers are biased towards the normal class, which means that fraud instances tend to be classified as normal ones. The imbalanced learning problem is a continous area of study \cite{Zhang2015}, which maybe solved via data sampling or cost-sensitive learning. 

\item Instead of using a single learning model, we will apply ensemble learning to build the optimal SB classification model in order to avoid the overfitting problem, lower the model's error ratio and decrease the bias and variance errors. The most popular ensemble methods are Bootstrap Aggregation (Bagging) and Boosting. Therefore, both methods will be investigated to determine the most effcient one four our SB dataset.  
     
\end{itemize}

\begin{figure} [H]
\centering
\includegraphics[scale=.7]{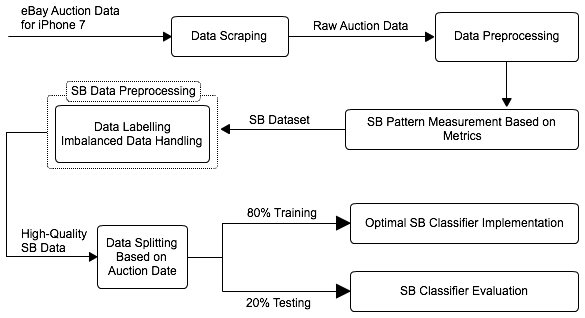} 
\caption{High-Quality SB Dataset Construction and Classification}
\end{figure}

\bibliographystyle{alpha}
\bibliography{References}
\end {document}